\documentclass[a4paper]{article}

\usepackage{INTERSPEECH2019}
\usepackage[utf8]{inputenc}
\usepackage{hyperref}

\name{Doug Beeferman, William Brannon, Deb Roy}
\address{Lab for Social Machines, MIT Media Lab}
\email{dougb5@mit.edu, wbrannon@mit.edu, dkroy@media.mit.edu}

\title{RadioTalk: a large-scale corpus of talk radio transcripts}

\begin{document}
\maketitle

\begin{abstract}
We introduce RadioTalk, a corpus of speech recognition transcripts sampled from talk radio broadcasts in the United States between October of 2018 and March of 2019.  The corpus is intended for use by researchers in the fields of natural language processing, conversational analysis, and the social sciences.
The corpus encompasses approximately 2.8 billion words of automatically transcribed speech from 284,000 hours of radio, together with metadata about the speech, such as geographical location, speaker turn boundaries, gender, and radio program information.  In this paper we summarize why and how we prepared the corpus, give some descriptive statistics on stations, shows and speakers, and carry out a few high-level analyses.
\end{abstract}

\section{Introduction}
Every day tens of thousands of conversations take place on American talk radio, a medium with enormous reach and influence. In 2017, fully 93\% of Americans age 18 and older listened to broadcast radio in a given week, and at any given time of day, news and talk stations commanded about 10\% of the total audience. \cite{pewresearch1}

Some of these conversations are local in scope, while others embrace national or international events. Some are in call-in shows that are syndicated across the country, while others are unique to a single location.

Radio is poorly studied relative to other parts of the public sphere such as social media and print and online news.  Radio listeners are disproportionately likely to be from demographics with low rates of social media use.  In particular, most older Americans are not Twitter users, with 19\% of those 50-64 and only 8\% of those 65 and older reporting use of Twitter in surveys. \cite{Pew2018} Radio, by contrast, reaches large numbers of older adults, with 92\% of those 50 and older listening to terrestrial radio in a given week. \cite{Nielsen2017} Because those calling in to radio shows are usually also listeners, a corpus of radio content is thus doubly useful: it captures an important form of media for these demographics, and the call-in content provides diverse examples of naturally occurring conversational speech.

Automatic conversational speech recognition is now fast enough for such a corpus to be practicable to collect at a large scale, and accurate enough to be useful for analysis.  In this paper we introduce a corpus of speech recognition transcripts sampled from talk radio broadcasts in the United States broadcast between October of 2018 and March of 2019, and we show how it can reveal insights relevant to conversation analysis, topic analysis, and the medium of talk radio itself.

\section{Related work}

Other corpora of conversational speech include the CALLHOME corpus \cite{callhomecorpus}, the Switchboard corpus \cite{godfrey1992switchboard} and the Fisher corpus \cite{cieri2004fisher}.  All of these emphasize telephone speech and include audio matched with transcripts.  Text-only corpora of discussions from online message boards are also widely available, such as the Reddit corpus released in 2015. \cite{redditcorpus}.

The authors are unaware of any corpora covering conversations on talk radio, although there are two widely cited data sets that focus more narrowly on news reports:  The Broadcast News corpus \cite{graff19971996} includes 130 hours of news on three television stations and one radio station;  and the Boston University Radio News Corpus \cite{ostendorf1995boston} includes 7 hours of speech read by news announcers from one radio station.  

Several researchers in the social sciences have analyzed smaller-scale sets of talk radio content, notably to measure the decline of local programming \cite{crider2012public}, to understand the power dynamics between talk show hosts and callers \cite{hutchby2013confrontation}, and to gauge and categorize incivility in public discourse \cite{sobieraj2011incivility}.  

\section{Corpus preparation}
The corpus discussed in this paper is the result of an ingestion and processing pipeline which we now briefly describe. This pipeline encompasses three stages, interacting with each other asynchronously through a data lake: ingestion of audio, transcription and post-processing.

\subsection{Ingestion}
The ingestion phase collects audio from online streams of radio stations which have made such streams publicly available on the Internet.  (See below for details on the included stations.) For greatest reliability, the ingestion processes run in separate, lightweight containers, writing the streamed audio to the data lake as they collect it. In the event of network difficulties, these processes reconnect and re-spawn as necessary to minimize downtime and avoid missing audio.

\subsection{Transcription}

The transcription system, which runs asynchronously with the ingestion, checks for new audio files and transcribes them, writing the transcripts back to the data lake. 

Our speech-to-text model is based on an entry by Peddinti et al. \cite{Aspire2015} in the IARPA ASpIRE challenge.  Its acoustic model has a time-delay neural network (TDNN) architecture geared for speech in reverberant environments, and offered an appropriate trade-off of accuracy on radio and decoding efficiency for our needs.  It is trained on the English portion of the Fisher corpus.

 To reduce word error rates, we replaced the lexicon and language model, retraining them on several corpora of human-transcribed radio: several years each of broadcasts from a conservative talk show \cite{Rush} and two National Public Radio news/talk shows.\cite{TOTN,ME}  Keeping current with these sources gives our system better coverage of proper names in the news.
 
 The final speech-to-text model is implemented with the commonly used Kaldi toolkit. \cite{Povey_ASRU2011}   We observed a word error rate of approximately 13.1\% with this system, as measured on a set of human-transcribed talk radio content that aired after the time period of the system's training data. \footnote{On the same basis, the Google Cloud Speech-to-Text API\cite{googlespeechapi} gave a 7.1\% word error rate, but its cost was prohibitive for the scale of our project, more than 40 times the cost per hour of our Kaldi-based solution.} 

\subsection{Post-processing}

The third step of processing appends other data generated from the audio, transcripts and station lists. These additional fields are intended to support use of the RadioTalk corpus for both NLP tasks and social science research on the radio ecosystem. Particularly important fields include:
\begin{itemize}
    \item Anonymous speaker identifiers and diarization (speaker turn boundaries)
    \item Confidence scores, the speech recognizer's estimate of its error rate aggregated at the speaker-turn level. 
    \item Imputed speaker gender
    \item A flag for whether a given utterance was recorded in a studio or came from a telephone call-in,
    \item Program/show identifiers and names, collected from scraped station schedules.  More than 1,300 unique shows were collected.
\end{itemize}

Speaker segmentation was performed using the LIUM speaker diarization toolkit \cite{meignier2010lium}, which uses spectral clustering to group audio sequences by speaker without supervision.   The gender and studio-vs-telephone classifiers were built within the same framework.

After these post-processing steps are performed, the content is cut into "snippets", or segments of speech from the same speaker turn.  An example of a record from the corpus is shown in Figure \ref{fig:example_snippet}.

\begin{figure}
  \centering
  
\begin{verbatim}
 "content": "Why are people dying more 
  often of opioid overdoses in the eastern
  part of the U.S compared to the western
  part what what do you think",
 "segment_start_time": 1543536684.69,
 "segment_end_time": 1543536692.95,
 "mean_word_confidence": 0.948,
 "speaker_id": "S2",
 "guessed_gender": "F",
 "studio_or_telephone": "S", 
 "callsign": "KNAG", 
 "city": "Grand Canyon",
 "state": "AZ",
 "show_name": "All Things Considered"
\end{verbatim}

  \caption{A single "snippet" record in the RadioTalk corpus.  Complete descriptions of these and other fields can be found in the corpus documentation.}
  \label{fig:example_snippet}
\end{figure}

\subsection{Radio Station Coverage}
Because all radio content airs on specific radio stations, the first problem in assembling a corpus like RadioTalk is choosing the set of stations to include. To enable a systematic selection process, we began by assembling a complete list of radio stations in the United States, together with various supplementary data. Most of our data was sourced from Radio-Locator \cite{RadioLocator}, a third-party company specializing in radio station data, with much of the data ultimately coming from federal regulatory filings. The Radio-Locator data provided a list of stations with call letters, postal addresses, a "format" variable indicating the type of programming the station airs, a URL for an online stream of the broadcast where available, and various other station-level variables.

This data set listed 17,124 stations, of which 1,912 were coded with talk or talk-related formats.\footnote{Specifically, "News", "Business News", "Farm", "Public Radio", "Talk", "College", and "News/Talk". Of these, 823 stations were Public Radio, and another 780 either Talk or News/Talk.} We considered these 1,912 stations the universe of talk-radio stations for inclusion in the sample. The initial list of stations to ingest and transcribe was a random sample of 50 stations from among this group, selected to be nationally representative and to permit weighting summary estimates back to the population of radio stations.\footnote{See the corpus website for the full details of the selection process.}

\begin{figure}
  \centering
    \includegraphics[scale=0.5]{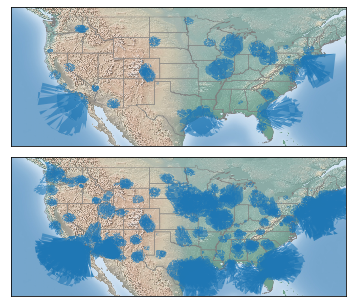}
  \caption{Total geographic reach, including water area, of the initial (top) and current (bottom) sets of transcribed radio stations.}
  \label{fig:coverage_map}
\end{figure}

After choosing and beginning to ingest this initial panel of stations, we added 242 other stations over the intervening months.\footnote{In an indication of the churn in radio stations, eight of the initial stations have ceased providing online streams or changed formats since the beginning of ingestion, and are not represented in later portions of the corpus.} These stations were not intended to be nationally representative, and focused on particular geographic areas of research interest to our team. Particularly large numbers of these later stations are in Wisconsin, near Lincoln, Nebraska, or in the Boston area. The initial and current panels of stations are shown in Figure \ref{fig:coverage_map}.

\section{Corpus overview}
In all, the corpus contains approximately 2.8 billion words of speech from 284,000 hours of radio between October 2018 and March 2019.  We sample 50\% of the utterances during the time period, including every other ten-minute period for each station.  This leaves large sections of dialogue intact for the analysis of conversational dynamics.

We assumed that all corpus content was in English for transcription purposes, as all selected stations primarily air English-language formats.\footnote{Rarely, there may be short periods of non-English speech in the underlying audio; if present, it should be represented as a sequence of "unknown" tokens.} To highlight the corpus's diverse content and wide range of applications, we present certain top-level analyses of the included stations, shows, and content.

\subsection{Speaker turn characteristics}
We can segment the corpus into \textit{speaker turns}, intervals of uninterrupted speech asserted to come from the same speaker.  Doing so yields 31.1 million speaker turns over the time period, which we can aggregate in various ways using the metadata fields provided by our pipeline.  Table \ref{table:1} shows some measures broken out by syndication level, studio/telephone voice, and gender.  

\begin{itemize}
\item Non-syndicated content makes up a about one-third (32.7\%) of the corpus, measured by number of speaker turns. Telephone speech makes up 10.0\% of the total, with similar representation in the local and syndicated subsets.
\item Female voices account for under one-third (27.8\%) of the content, with similar representation in the local and syndicated subsets.  Female voices account for a substantially larger share of the telephone subset (32.6\%) than of the studio subset (27.3\%), suggesting that call-in voices are more gender-balanced than talk show hosts.
\item Speaker turns are 13\% longer in the telephone subset than in the studio subset, and 21.9\% longer for male speech than for female speech. \footnote{While these differences are dramatic, we should caution that we haven't evaluated the diarization and gender classification pipeline sufficiently to be certain that its errors aren't correlated in ways that could distort these numbers.}
\item The confidence score aggregates suggest that the recognizer has a harder time with telephone speech than studio speech, making relatively (9.3\%) more word errors; and a harder time with female speech than male speech, making relatively (7.4\%) more word errors.

\end{itemize}

\begin{table}
\centering
\begin{tabular}{| p{2.0em} | p{3em} | p{3.0em}| p{3.5em} | p{4em} | p{2.5em} |}
 \hline
 Synd. & Studio / Phone & Gender & Fraction of corpus & Mean Duration (sec) & Mean reco. conf. \\
\hline\hline
 Yes & Studio & Female & 16.8\% & 15.43 & 0.874 \\ 
\hline
 Yes & Studio & Male & 43.7\% & 18.04 & 0.885 \\ 
\hline
 Yes & Phone & Female & 2.2\% & 15.43 & 0.862 \\ 
\hline
 Yes & Phone & Male & 4.6\% & 22.41 & 0.874 \\ 
\hline\hline
 No & Studio & Female & 7.9\% & 14.16 & 0.854 \\ 
\hline
 No & Studio & Male & 21.7\% & 17.65 & 0.867 \\ 
\hline
 No & Phone & Female & 1.1\% & 13.62 & 0.846 \\ 
\hline
 No & Phone & Male & 2.1\% & 19.98 & 0.860 \\

\hline
\end{tabular}
\caption{Some properties of the 31.1 million speaker turns in the RadioTalk corpus. "Synd." refers to whether the turn comes from a radio show which is known to be syndicated across multiple stations. "Mean reco. conf." refers to the mean speech recognizer confidence score for the subset, an estimate of the fraction of correctly transcribed words.}
\label{table:1}
\end{table}

\subsection{Topics discussed}
The lexical composition of a corpus of radio transcripts will naturally reflect the interests and perspectives of the people whose voices are in the news programs and call-in shows that it captures.   These interests are an amalgam of topics of local, national, and international concern.  The period of October to December, 2018, was particularly rich with national news related to the US general election near its midpoint, November 7.

\begin{figure}
  \centering
    \includegraphics[scale=0.55]{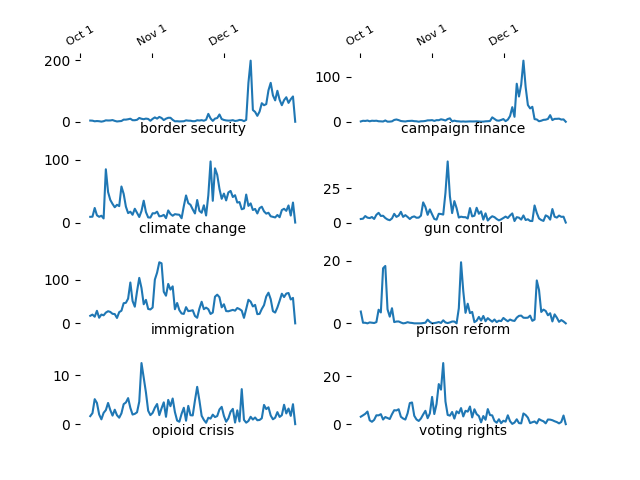}
  \caption{Time series charts showing the relative number of mentions per day for several phrases related to issues discussed on talk radio between October and December of 2018.  The y-axis for each chart is the number of mentions of the phrase per million words transcribed that day.}
  \label{fig:issue_mentions}
\end{figure}

Figure \ref{fig:issue_mentions} gives a glimpse of eight topics that were top of mind during this quarter.   Discussion of immigration policy and voting rights peaked leading up to the election.  The term "border security" gained currency in late December in reference to a proposed border wall.  Gun control discussion spiked after mass shootings in October and November, while interest in climate change tracked major weather events such as hurricanes and wildfires.  

\begin{figure}
  \centering
    \includegraphics[scale=0.52]{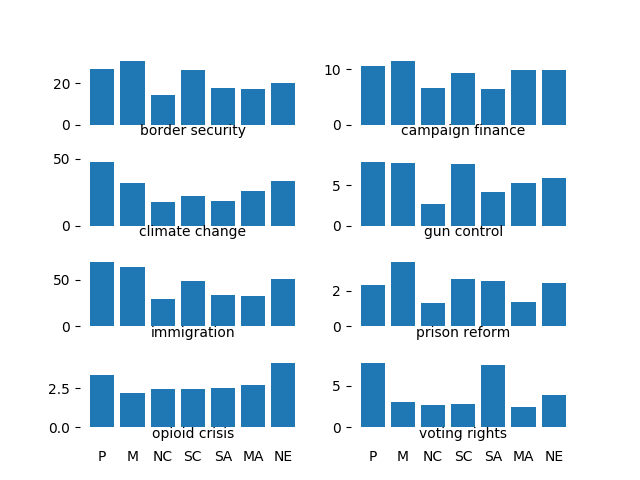}
  \caption{The relative number of mentions per day for the same phrases as in Figure \ref{fig:issue_mentions}, but organized by sub-region of the United States.  Here again, the y-axis for each chart is the number of mentions of the phrase per million words transcribed in the region.  The x-axis is the census sub-region, listed west to east:  P: Pacific (AK, CA, OR, WA);  M: Mountain (AZ, CO, ID, MT, NM, NV, UT);  NC: North Central (IA, MI, MN, MO, ND, NE, OH, SD, WI);  SC: South Central (AL, MS, TN, TX);  SA: South Atlantic (DC, FL, GA, SC, VA, WV);  MA:  Mid Atlantic (NY, PA);  NE: New England (CT, MA, ME, NH, RI)}
  \label{fig:issue_geos}
\end{figure}

Figure \ref{fig:issue_geos} shows the same mentions grouped by geographical sub-region of the United States.  For example, climate change is more frequently discussed on the coasts, opioid discussion has the largest share of voice in New England, and voting rights was frequently discussed in the South Atlantic region, where Florida voters approved a constitutional amendment restoring voting rights to people with past felony convictions.

\subsection{Radio programs}

Another way to cut the data is by radio program, using the show identifier inferred for each
record based on publicly available station schedule data.  Table \ref{table:radioshows} shows selected
properties of the most widely syndicated radio shows in the corpus, which include a variety
of nationally talk shows and news programs.  General-interest news shows such as \textit{Morning Edition} have
the greatest lexical diversity, while more narrowly scoped programs like \textit{Marketplace}, a business news show,
have the least.  Talk shows have the greatest fraction of telephone speech and also the briskest
conversations as measured by the amount of silence between speaker turns.

\begin{table}
\centering
\begin{tabular}{| p{7em} | p{2em}| p{3.5em} | p{3.5em} | p{3em} |}
 \hline
 Show name & \# & Percent call-in speech & Lexical diversity & Inter-speaker silence (sec) \\
 \hline\hline
 Coast to Coast AM with George Noory & 48 & 44.2\% & 402 & 0.570 \\
 \hline
 The Sean Hannity Show & 47 & 11.1\% & 413 & 0.560 \\
 \hline
 Rush Limbaugh & 46 & 6.4\% & 428 & 0.426 \\
 \hline
 All Things Considered & 42 & 3.4\% & 451 & 0.465 \\
 \hline
 Morning Edition & 42 & 5.2\% & 453 & 0.571 \\
 \hline
 Fresh Air & 41 & 3.3\% & 408 & 0.622 \\
 \hline
 This American Life & 39 & 4.6\% & 415 & 1.03 \\
 \hline
 1A & 36 & 3.1\% & 421 & 0.501 \\
 \hline
 BBC World Service & 36 & 2.3\% & 437 & 0.513 \\
 \hline
 Marketplace & 35 & 3.1\% & 392 & 0.795 \\
 \hline
 \end{tabular}
\caption{Some properties of the top 10 most widely
syndicated radio shows observed in the corpus.
The second column gives the number of stations in the corpus
which air the show.  For this summary, a single airing for each episode
was selected from the corpus based on recognizer confidence.
"Lexical diversity" refers to the mean number of unique words
seen in any window of 1000 words \cite{covington2010cutting}}

\label{table:radioshows}
\end{table}

\subsection{Syndication network}
We can also consider the network formed between the stations by the syndicated content they air. In this undirected syndication network, two stations are connected if they air any of the same programs.\footnote{Note that syndication is not necessarily real-time, and these programs need not air simultaneously or for the same length of time.}

Network analysis is a rich and fruitful way of analyzing station relationships, but for brevity we only summarize the syndication network here. Of the the 183 stations with schedule data, one has no syndication links to other stations. The remaining 182 are connected by 6,736 edges, forming a single connected component with an average degree of 73.

\begin{figure}[!ht]
  \centering
    \includegraphics[scale=0.23]{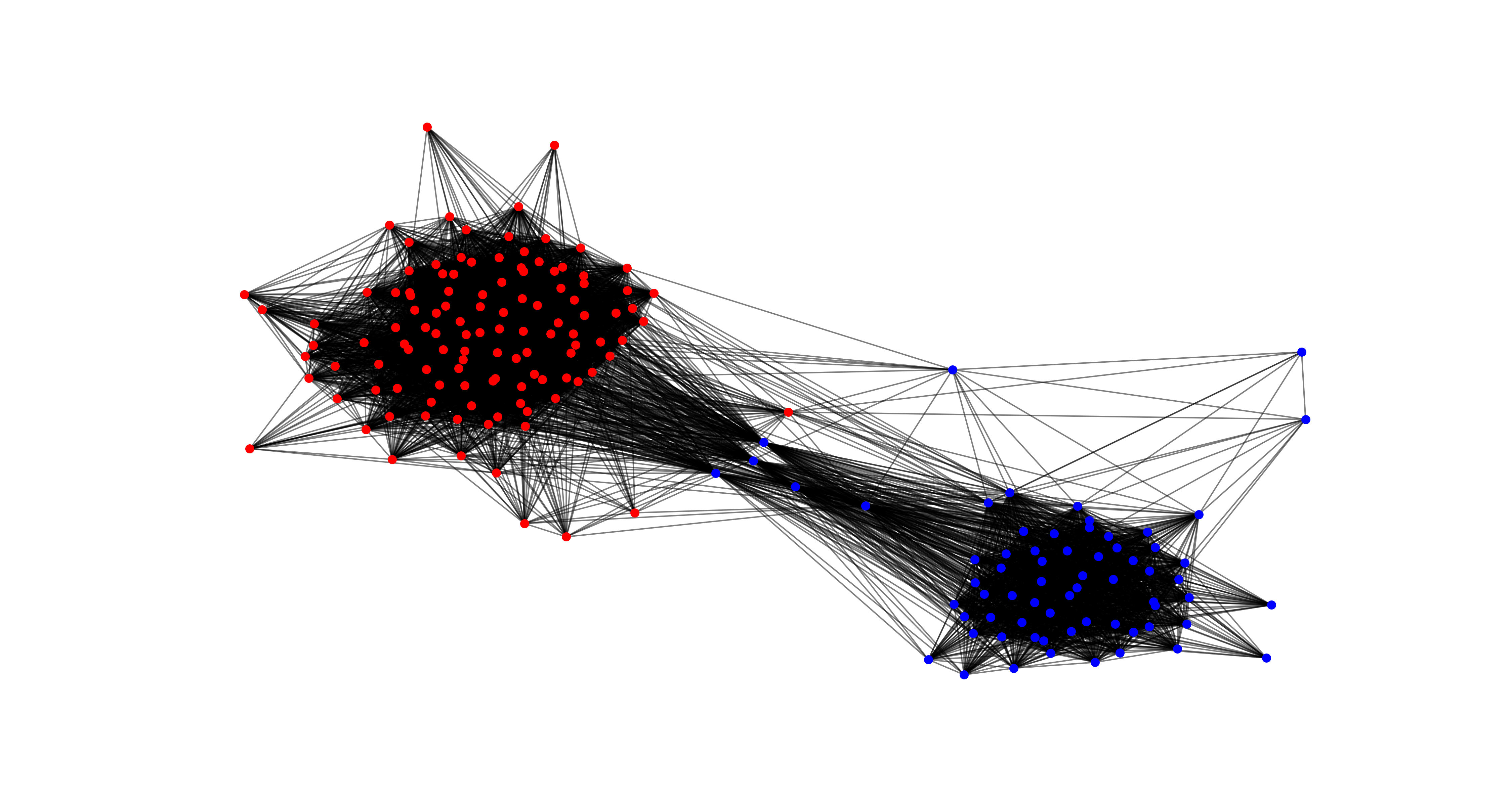}
  \caption{The syndication network between the stations, with any two stations connected if there are any programs airing on both. The larger, conservative radio station community is colored red, and the smaller, liberal or public-radio community is in blue. One station with degree 0 is not shown.}
  \label{fig:syndication_network}
\end{figure}

The Louvain algorithm for community detection \cite{louvain} identifies two communities in the network, of sizes 116 and 67. Manual inspection suggests that the larger community represents conservative talk radio stations, and the smaller one liberal or public radio stations. The network with these communities color-coded is displayed in Figure \ref{fig:syndication_network}.

\section{Conclusion}
This paper introduces RadioTalk, a corpus of transcribed speech broadcast on talk radio stations throughout the United States. The corpus includes scripted and conversational speech from a large and diverse set of speakers, and includes speaker-, program- and station-level metadata. Despite the presence of transcription error, RadioTalk shows promise for a wide range of questions in social science and natural language processing.

More information on the RadioTalk corpus is available at \href{https://github.com/social-machines/RadioTalk}{https://github.com/social-machines/RadioTalk}.  New versions may be released in the future with additional transcribed audio, improved transcriptions of the current corpus, or additional fields derived from the audio.

\section{Acknowledgements}
The authors would like to thank the Ethics and Governance of AI Fund for supporting this work. We also thank Brandon Roy, Clara Vandeweerdt, Wes Chow, Allison King, and the many colleagues at Cortico and the Laboratory for Social Machines who provided suggestions and feedback.

\bibliographystyle{IEEEtran}
\bibliography{main}

\end{document}